\documentclass{article}
\usepackage[utf8]{inputenc}
\usepackage{a4wide}
\usepackage{xcolor}
\usepackage{colortbl}
\usepackage{graphicx}
\usepackage{longtable}
\usepackage{multirow}

\begin{document}

\title{Comparison of classifiers in challenge scheme}
\author{Sergio Nava-Muñoz$^{1}$ and Mario Graff Guerrero$^{2}$ and Hugo Jair Escalante$^{3}$}

\date{\small{$^{1}$ CIMAT, Aguascalientes, Mexico} ~\\
\small{$^{2}$ INFOTEC, Aguascalientes, Mexico}~\\
\small{$^{3}$ INAOE, Puebla, Mexico}~\\
nava@cimat.mx, mario.graff@infotec.mx, hugojair@inaoep.mx~\\
This work has been accepted for publication in the~\\
Mexican Conference on Pattern Recognition~\\
MCPR2023}

\maketitle

\begin{abstract}
   In recent decades, challenges have become very popular in scientific research as these are crowdsourcing schemes. In particular, challenges are essential for developing machine learning algorithms. For the challenges settings, it is vital to establish the scientific question, the dataset (with adequate quality, quantity, diversity, and complexity), performance metrics, as well as a way to authenticate the participants' results (\textit{Gold Standard}). This paper addresses the problem of evaluating the performance of different competitors (algorithms) under the restrictions imposed by the challenge scheme, such as the comparison of multiple competitors with a unique \textit{dataset} (with fixed size), a minimal number of submissions and, a set of metrics chosen to assess performance. The algorithms are sorted according to the performance metric. Still, it is common to observe performance differences among competitors as small as hundredths or even thousandths, so the question is whether the differences are significant. This paper analyzes the results of the \textit{MeOffendEs@IberLEF 2021} competition and proposes to make inference through resampling techniques (bootstrap) to support Challenge organizers' decision-making.  
\end{abstract}

%In this paper, we show which competitors achieved the highest performance and those who performed the same.
\section{Introduction}

A challenge is a collaborative competition that has gained appeal among research scientists during the past couple decades. This challenges makes use of leaderboards so that participants may keep track of how they compare to other participants in terms of performance.  These cover multiple areas of science and technology, ranging from fundamental to applied questions in machine learning (kaggle, codalab \cite{Pavao2022}). At the most basic level, a performance benchmark requires a task, a metric, and a means of authenticating the result. 

Participants in this type of crowdsourcing are given data together with the specific question that needs to be answered. The creators of such challenges have ``ground truth'' or ``Gold Standard'' data that is exclusive to them and enables them to impartially rate the techniques that competitors create.
Participants provide their solutions for evaluation by the organizers using the Gold Standard data. In this way, it is possible to ﬁnd the best available method to solve the problem posed, and the participants can get an objective assessment of their methods. Clear scoring mechanisms for solutions evaluation and the availability of non-public datasets for use as Gold Standards are necessary for the organization of these challenges.

If we talk about a challenge task where the performance of classification algorithms is compared, there are particular constraints, such as the comparison of multiple participants (algorithms, methods, or assembles), selected performance metrics, a fixed dataset size, and a limit number of submissions per participant. It is difficult to use classical statistics to infer the significance of a given performance metric because there are no multiple datasets or many submissions; besides, our interest is making multiple comparisons.   

This research aims to complement the winner selection process based solely on the score rank by proposing the use of well-established statistical tools and error bar plots to facilitate the analysis of the performance differences and being able to identify whether the difference in performance is significant or the result of chance. To illustrate the approach, the dataset of subtasks 3 of \textit{MeOffendEs@IberLEF 2021} \cite{Plaza-del-Arco2021} competition is used. The analysis shows that for each performance measure, there is a system that performs the best; however, in precision, three other systems behave similarly; for the $F_1$ score, there are two systems with similar performance, and for recall, there are no similar systems. 

%% It specifically made use of the OffendMEX dataset, a novel collection of Mexican Spanish tweets that were manually labeled as offensive and obtained via Twitter.

The remainder of this article is organized as follows. Collaborative competitions are described in Section \ref{sec:collaborative}. The dataset used to test this proposal is described in Section \ref{sec:dataset}. The proposed solutions to the problem are described in Section \ref{sec:approach}, and their results are evaluated. In Section \ref{sec:conclusions}, conclusions are presented.

\section{Collaborative competitions}
\label{sec:collaborative}

\textit{Crowdsourcing} is a term introduced in 2006 by Jeff Howe, journalist, and editor of the electronic magazine Wired; however, it is a way of working that has been used for centuries. This practice allows a task to be performed by a set of experts or not, through a call for proposals. Crowdsourcing has been used in various areas, for example, in marketing, astronomy, and genetics. But our interest is in the scientific field, where the idea is to call on the community to solve a scientific problem. One of the forms of crowdsourcing (collaborative work) that has taken great importance in recent years, is collaborative competitions also called \textit{challenges}, and in particular, we are interested in applications in science and technology. 

In the scientific field, crowdsourcing and benchmarking have been combined, leading to the development of solutions that quickly outperform the state-of-the-art. The essential elements of a challenge are the scientific question in the form of a task, the data (a single dataset of fixed size), the performance metrics, and a way to verify the results (the gold standard). Nonetheless, these characteristics limit the use of classical statistics to infer the significance of a given performance metric because there are no multiple datasets or many submissions.

\section{Dataset - \textit{MeOffendEs@IberLEF 2021}}
\label{sec:dataset}

This paper analyze MeOﬀendES 2021 dataset, organized at IberLEF 2021 and co-located with the 37th International Conference of the Spanish Society for Natural Language Processing (SEPLN 2021). MeOffendEs' major objective is to advance research into the recognition of offensive language in Spanish-language variants. The shared task consists of four subtasks. The first two relate to identifying offensive language categories in texts written in generic Spanish on various social media platforms, while subtasks 3 and 4 are concerned with identifying offensive language that is directed at the Mexican variant of Spanish. In particular, the focus is on Subtask 3: Mexican Spanish non-contextual binary classification. Participants must classify tweets in the OffendMEX corpus as offensive or non-offensive. For evaluation, we consider the offensive class's precision, recall, and f1 score\cite{Plaza-del-Arco2021}.  OffendMEX corpus, is a novel collection of Mexican Spanish tweets that were manually labeled as offensive and obtained via Twitter

The dataset used for this analysis is the \textit{test partition} from OffendMEX, for subtask three at \textit{MeOffendEs@IberLEF 2021}; this consists of $11$ variables, which correspond to the predictions of $10$ teams and the \textit{gold standard}. One of the competitors (NLPCIC) submitted a prediction after the competition had ended, and the system recorded it. The \textit{gold standard}  contains the labels for $600$ offensive tweets and $1582$ non-offensive tweets,  for a total of $n=2182$ tweets.

Table \ref{tab:metricas} summarizes the results using in terms of Precision, Recall, and $F_1$ scores. As can be seen, the highest $F_1$ score is $0.7154$ achieved by $NLPCIC$, followed by $CIMATMTYGTO$ with $0.7026$ and $DCCDINFOTEC$ with $0.6847$.  For Recall, the three best teams were $CENAmrita$, $xjywing$ and $aomar$ with values of $0.9183$, $0.8883$, and $0.8750$, respectively. Regarding Precision, $NLPCIC$ obtained the highest score with $0.7208$. $DCCDINFOTEC$ and $CIMATGTO$ came in second and third, with $0.6966$ and $0.6958$, respectively.

\begin{table}[!h]
\caption{\label{tab:metricas}Results for the Non-contextual binary classification for Mexican Spanish}
\centering
\begin{tabular}[t]{lrrc}
\hline
Team & precision & recall & $F_1$\\
\hline
\cellcolor{gray!6}{NLPCIC} & \cellcolor{gray!6}{0.7208} & \cellcolor{gray!6}{0.7100} & \cellcolor{gray!6}{0.7154}\\
CIMATMTYGTO & 0.6533 & 0.7600 & 0.7026\\
\cellcolor{gray!6}{DCCDINFOTEC} & \cellcolor{gray!6}{0.6966} & \cellcolor{gray!6}{0.6733} & \cellcolor{gray!6}{0.6847}\\
CIMATGTO & 0.6958 & 0.6633 & 0.6792\\
\cellcolor{gray!6}{UMUTeam} & \cellcolor{gray!6}{0.6763} & \cellcolor{gray!6}{0.6650} & \cellcolor{gray!6}{0.6706}\\
Timen & 0.6081 & 0.6000 & 0.6040\\
\cellcolor{gray!6}{CICIPN} & \cellcolor{gray!6}{0.6874} & \cellcolor{gray!6}{0.5350} & \cellcolor{gray!6}{0.6017}\\
xjywing & 0.3419 & 0.8883 & 0.4937\\
\cellcolor{gray!6}{aomar} & \cellcolor{gray!6}{0.3241} & \cellcolor{gray!6}{0.8750} & \cellcolor{gray!6}{0.4730}\\
CENAmrita & 0.3145 & 0.9183 & 0.4685\\
\hline
\end{tabular}
\end{table}

\section{Proposed approaches and results}
\label{sec:approach}

As mentioned, the objective is to propose tools that allow comparing the classification results of different competitors in a Challenge, in addition to performance metrics. In the literature, one can find works dealing with the problem of comparing classification algorithms; however, they focus on something other than the competition scheme. For example, Diettrich (1998) \cite{Dietterich1998}, reviews five proximate statistical tests to determine whether one learning algorithm outperforms another in a particular learning task. However, it is required to have the algorithm, and in our case we only have the prediction, not the algorithm. On the other hand, Dem{\v{s}}ar (2006) \cite{Demsar2006} focuses on Statistical Comparisons of Classifiers over Multiple Data Sets; however, in our case we have only one dataset. In particular, he presents several non-parametric methods, and several guides to performing a correct analysis when comparing a set of classifiers. García and Herrera (2008) \cite{Garcia2008} attack a problem similar to Dem{\v{s}}ar but focused on pairwise comparisons, i.e. statistical procedures for comparing $c \times c$ classifiers, but again on multiple datasets.

\subsection{Bootstrap}
\label{sec:bootstrap}

The word ``bootstrapping'' in statistics refers to drawing conclusions about a statistics' sampling distribution by resampling the sample with replacement data as though it were a population with a fixed size \cite{Chernick2011,Efron1994}. The term resampling was originally used in 1935 by R. A. Fisher in his famous randomization test and in 1937 and 1938 by E. J. G. Pitman, but in these instances the sampling was carried out without replacement.

The theory and applications of the bootstrap have exploded in recent years, and the Monte Carlo approximation to the bootstrap has developed into a well-established method for drawing statistical conclusions without making firm parametric assumptions. The term bootstrap refers to a variety of methods that are now included under the broad category of nonparametric statistics known as resampling methods. Brad Efron's publication in the \textit{Annals of Statistics} was published in 1979, making it a crucial year for the bootstrap \cite{Efron1979,Efron1994}. The bootstrap resampling technique was developed by Efron. His initial objective was to extract features of the bootstrap in order to better understand the jackknife (an earlier resampling technique created by John Tukey). He built it as a straightforward approximation to that technique. However, as a resampling method, the bootstrap frequently performs as well as or better than the jackknife.

Bootstrap has already been applied in NLP, it has been applied in the analysis of statistical significance in NLP systems. For instance, in the study conducted by Koehn (2004) \cite{Koehn2004}, bootstrap was used to estimate the statistical significance of the BLEU score in Machine Translation (MT). Likewise, in the research conducted by Zhang (2004) \cite{Zhang2004}, it was employed to measure the confidence intervals for BLEU/NIST score. Additionally, in the field of automatic speech recognition (ARS), researchers have used bootstrap to estimate confidence intervals in performance evaluation, as illustrated in the work of Bisani (2004) \cite{Bisani2004}. Despite the fact that using bootstrap in NLP problems is not a novel technique, it remains highly relevant.

\subsection{Comparison of Classifiers}
\label{sec:comparison1}

%% \cite{Raschka2018}

Comparing classification algorithms is a complex and ongoing problem. Performance can be defined in many ways, such as accuracy, speed, cost, etc., but accuracy is the most commonly used performance indicator. There are numerous accuracy measures that have been presented in the classification literature, with some specifically designed to compare classifiers and others originally defined for other purposes \cite{Labatut2012}. We will focus on challenge accuracy measures, see Table \ref{tab:metricas}.

The main objective is to make inferences on the performance parameter $\theta$ of the algorithms developed by the teams participating in the competition. This inference is made on a single dataset of size $n$, with a minimal amount of submissions. The inference concerns the parameter's value (performance) in the population from which the dataset is considered to be randomly drawn.

There are two classical approaches for making parameter inferences, hypothesis testing, and confidence intervals; in this contribution, inference on performance is made using both approaches, specifically bootstrap estimates. The procedure consists of extracting 10,000 Bootstrap samples (with replacement of size $n$) from the data set that includes the $n$ gold standard examples and the corresponding predictions. Each team's performance parameters are calculated for each sample, and the sampling distribution is obtained. Using the sampling distribution, the $95\%$ confidence interval for the mean of the performance parameter is constructed. Table  \ref{tab:tabla-boot-media}  contains the ordered estimates of the mean and confidence interval obtained through Bootstrap.  

\begin{table}[!h]
\caption{\label{tab:tabla-boot-media}Ordered Bootstrap Confidence Intervals}
\centering
\resizebox{\linewidth}{!}{
\fontsize{7}{9}\selectfont
\begin{tabular}[t]{lclclc}
\hline
\multicolumn{2}{c}{Precision} & \multicolumn{2}{c}{Recall} & \multicolumn{2}{c}{$F_1$} \\
Team & CI & Team & CI & Team & CI\\
\hline
\cellcolor{gray!6}{NLPCIC} & \cellcolor{gray!6}{(0.6844,0.7572)} & \cellcolor{gray!6}{CENAmrita} & \cellcolor{gray!6}{(0.8962,0.9402)} & \cellcolor{gray!6}{NLPCIC} & \cellcolor{gray!6}{(0.6864,0.7438)}\\
DCCDINFOTEC & (0.6585,0.7345) & xjywing & (0.8632,0.9134) & CIMATMTYGTO & (0.6739,0.7306)\\
\cellcolor{gray!6}{CIMATGTO} & \cellcolor{gray!6}{(0.6578,0.7338)} & \cellcolor{gray!6}{aomar} & \cellcolor{gray!6}{(0.8485,0.9015)} & \cellcolor{gray!6}{DCCDINFOTEC} & \cellcolor{gray!6}{(0.6536,0.7152)}\\
CICIPN & (0.6458,0.7290) & CIMATMTYGTO & (0.7260,0.7935) & CIMATGTO & (0.6481,0.7098)\\
\cellcolor{gray!6}{UMUTeam} & \cellcolor{gray!6}{(0.6381,0.7143)} & \cellcolor{gray!6}{NLPCIC} & \cellcolor{gray!6}{(0.6739,0.7458)} & \cellcolor{gray!6}{UMUTeam} & \cellcolor{gray!6}{(0.6393,0.7011)}\\
CIMATMTYGTO & (0.6175,0.6888) & DCCDINFOTEC & (0.6351,0.7112) & Timen & (0.5713,0.6365)\\
\cellcolor{gray!6}{Timen} & \cellcolor{gray!6}{(0.5691,0.6474)} & \cellcolor{gray!6}{UMUTeam} & \cellcolor{gray!6}{(0.6269,0.7025)} & \cellcolor{gray!6}{CICIPN} & \cellcolor{gray!6}{(0.5665,0.6363)}\\
xjywing & (0.3182,0.3656) & CIMATGTO & (0.6255,0.7011) & xjywing & (0.4676,0.5196)\\
\cellcolor{gray!6}{aomar} & \cellcolor{gray!6}{(0.3011,0.3470)} & \cellcolor{gray!6}{Timen} & \cellcolor{gray!6}{(0.5608,0.6392)} & \cellcolor{gray!6}{aomar} & \cellcolor{gray!6}{(0.4470,0.4987)}\\
CENAmrita & (0.2926,0.3364) & CICIPN & (0.4946,0.5751) & CENAmrita & (0.4433,0.4935)\\
\hline
\end{tabular}}
\end{table}

\subsection{Comparison of Classifiers through Independents samples}
\label{sec:independents}

\begin{figure}[ht]
\centering
\includegraphics[width=1\textwidth]{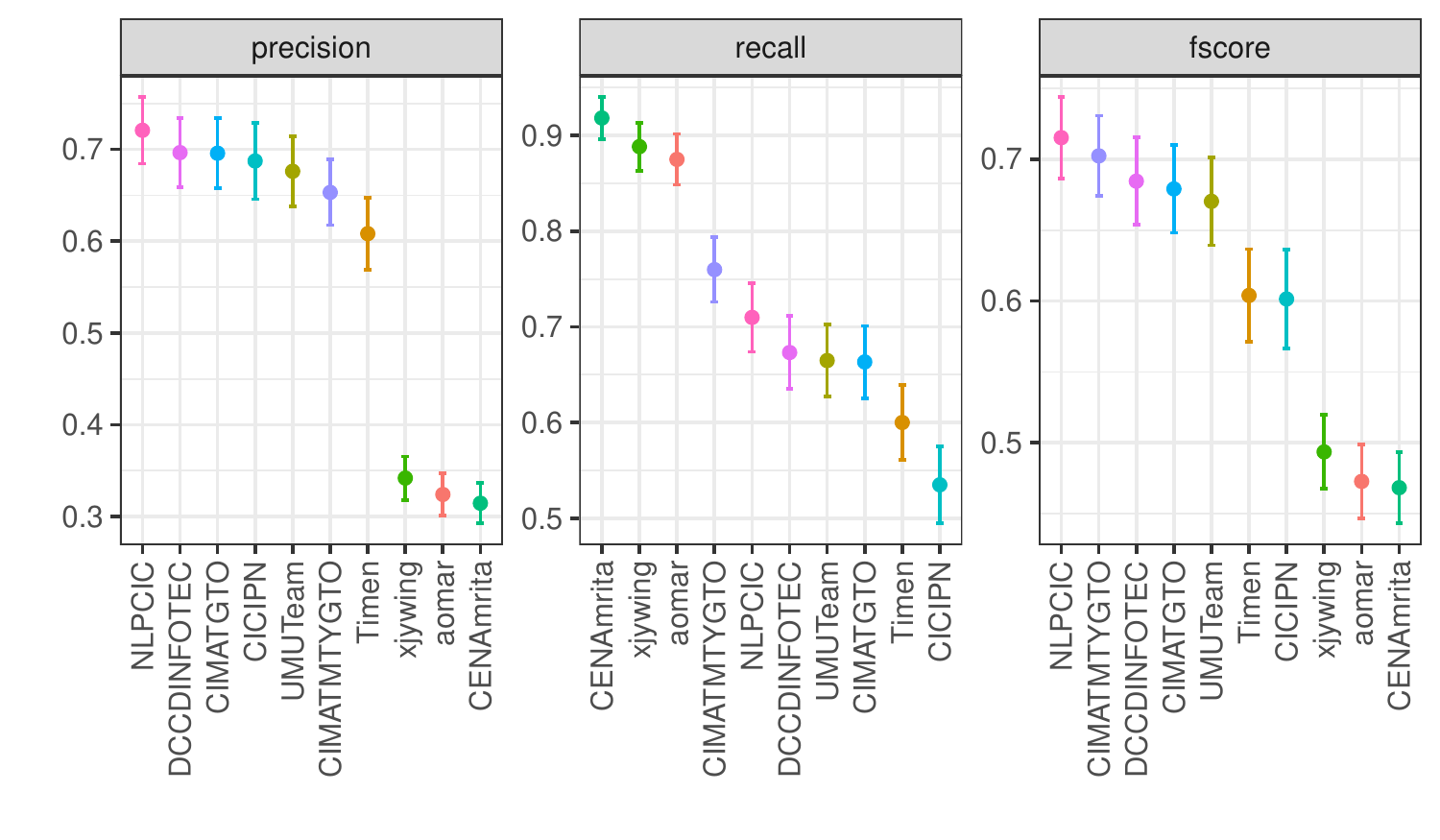}
\caption{Ordered Bootstrap Confidence Intervals}
\label{fig:grafica1}
\end{figure}

Suppose the confidence intervals of the means of two populations overlap; in that case, this is enough to conclude that there is no significant difference between the means of the populations. On the other hand, if the intervals do not overlap, then there is an indication that the difference in performance is significant. The hypothesis testing approach would be set the null hypothesis $H_0$,  that $\theta_i = \theta_j$; against the alternative hypothesis $H_1$ that  $\theta_i \neq \theta_j$, for $i \neq j$. In the case of the performance metrics of the algorithms, the confidence intervals at $95\%$ can be observed in Table \ref{tab:tabla-boot-media} and Figure \ref{fig:grafica1}. Intervals have been ordered to facilitate interpretation. As can be seen, the team with the highest $F_1$ score is $NLPCIC$, with $95\%$ confidence interval equal to $(0.6864,0.7438)$. The second place corresponds to $CIMATMTYGTO$ $(0.6739,0.7306)$. As the first two intervals overlap, it suggests that the $F_1$ scores of both teams will likely be the same in the population from which the dataset was sampled. Conversely, there is a difference between $NLPCIC$ and $Timen$.

\subsection{Comparison of classifiers through paired samples}
\label{sec:paired}

\begin{figure}[ht]
\centering
\includegraphics[width=1\textwidth]{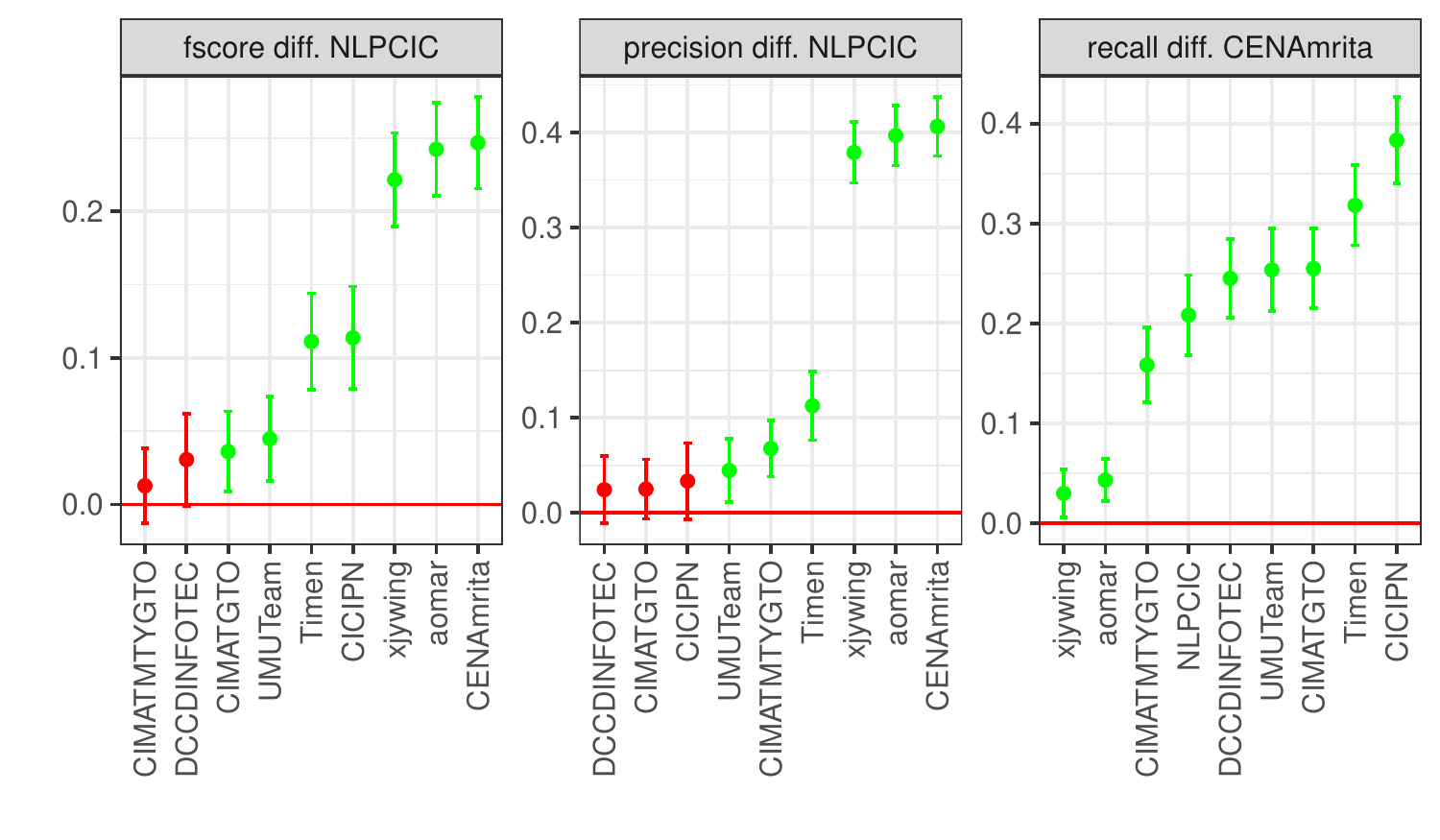}
\caption{Bootstrap Confidence Intervals of differences with the best-performing competitor. Red intervals contain zero, and green intervals do not contain it.}
\label{fig:grafica2}
\end{figure}

However, since each Bootstrap sample contains both the gold standard and each team's prediction for the same tweet, it is possible to calculate the performance and also the performance difference for each sample, i.e., the paired bootstrap method is being used \cite{Chernick2011,Efron1994}. Confidence intervals at the $95\%$ level for the difference in performance between paired samples were constructed, following the same approach as in the previous case. Table \ref{tab:tabla-ref-media-2} and Figure \ref{fig:grafica2} display the confidence intervals for comparing the top team with the others. In this case, for the difference, if the interval contains zero, it is impossible to rule out that the performance of both algorithms is the same in the population from which the dataset was obtained. In other words, $H_0$ cannot be rejected. For the $F_1$ score, the team with the highest performance is $NLPCIC$, and as can be seen, we cannot rule out that its performance is the same as $CIMATMTYGTO$ and $DCCDINFOTEC$. On the other hand, there are significant differences compared to the rest of the teams using $F_1$ score. Regarding recall, the team with the best performance was $CENAmrita$, and no other team had the same performance. Concerning precision, the team with the best performance is $NLPCIC$; however, we cannot rule out that  $DCCDINFOTEC$, $CIMATGTO$, and $CICIPN$ have equivalent performances.

\begin{table}[h]

\caption{\label{tab:tabla-ref-media-2}Bootstrap Confidence Intervals of differences from the best.}
\centering
\resizebox{\linewidth}{!}{
\fontsize{7}{9}\selectfont
\begin{tabular}[t]{>{}lrrrlrrrlrrr}
\hline
\multicolumn{4}{c}{Precision} & \multicolumn{4}{c}{Recall} & \multicolumn{4}{c}{$F_1$} \\
\multicolumn{4}{c}{NLPCIC} & \multicolumn{4}{c}{CENAmrita} & \multicolumn{4}{c}{NLPCIC} \\
Team & ICI & Mean & SCI & Team & ICI & Mean & SCI & Team & ICI & Mean & SCI\\
\hline
\cellcolor{gray!6}{DCCDINFOTEC} & \cellcolor{gray!6}{-0.0110} & \cellcolor{gray!6}{0.0243} & \cellcolor{gray!6}{0.0596} & \cellcolor{gray!6}{xjywing} & \cellcolor{gray!6}{0.0060} & \cellcolor{gray!6}{0.0299} & \cellcolor{gray!6}{0.0539} & \cellcolor{gray!6}{CIMATMTYGTO} & \cellcolor{gray!6}{-0.0128} & \cellcolor{gray!6}{0.0128} & \cellcolor{gray!6}{0.0385}\\
CIMATGTO & -0.0063 & 0.0250 & 0.0563 & aomar & 0.0221 & 0.0432 & 0.0643 & DCCDINFOTEC & -0.0008 & 0.0307 & 0.0621\\
\cellcolor{gray!6}{CICIPN} & \cellcolor{gray!6}{-0.0065} & \cellcolor{gray!6}{0.0334} & \cellcolor{gray!6}{0.0733} & \cellcolor{gray!6}{CIMATMTYGTO} & \cellcolor{gray!6}{0.1211} & \cellcolor{gray!6}{0.1585} & \cellcolor{gray!6}{0.1958} & \cellcolor{gray!6}{CIMATGTO} & \cellcolor{gray!6}{0.0087} & \cellcolor{gray!6}{0.0361} & \cellcolor{gray!6}{0.0635}\\
UMUTeam & 0.0116 & 0.0446 & 0.0776 & NLPCIC & 0.1683 & 0.2084 & 0.2485 & UMUTeam & 0.0161 & 0.0449 & 0.0736\\
\cellcolor{gray!6}{CIMATMTYGTO} & \cellcolor{gray!6}{0.0380} & \cellcolor{gray!6}{0.0677} & \cellcolor{gray!6}{0.0974} & \cellcolor{gray!6}{DCCDINFOTEC} & \cellcolor{gray!6}{0.2058} & \cellcolor{gray!6}{0.2451} & \cellcolor{gray!6}{0.2844} & \cellcolor{gray!6}{Timen} & \cellcolor{gray!6}{0.0784} & \cellcolor{gray!6}{0.1112} & \cellcolor{gray!6}{0.1440}\\
Timen & 0.0763 & 0.1126 & 0.1488 & UMUTeam & 0.2122 & 0.2535 & 0.2948 & CICIPN & 0.0788 & 0.1137 & 0.1486\\
\cellcolor{gray!6}{xjywing} & \cellcolor{gray!6}{0.3471} & \cellcolor{gray!6}{0.3789} & \cellcolor{gray!6}{0.4108} & \cellcolor{gray!6}{CIMATGTO} & \cellcolor{gray!6}{0.2150} & \cellcolor{gray!6}{0.2549} & \cellcolor{gray!6}{0.2949} & \cellcolor{gray!6}{xjywing} & \cellcolor{gray!6}{0.1896} & \cellcolor{gray!6}{0.2215} & \cellcolor{gray!6}{0.2534}\\
aomar & 0.3651 & 0.3967 & 0.4284 & Timen & 0.2782 & 0.3182 & 0.3582 & aomar & 0.2105 & 0.2422 & 0.2740\\
\cellcolor{gray!6}{CENAmrita} & \cellcolor{gray!6}{0.3753} & \cellcolor{gray!6}{0.4063} & \cellcolor{gray!6}{0.4373} & \cellcolor{gray!6}{CICIPN} & \cellcolor{gray!6}{0.3401} & \cellcolor{gray!6}{0.3833} & \cellcolor{gray!6}{0.4266} & \cellcolor{gray!6}{CENAmrita} & \cellcolor{gray!6}{0.2155} & \cellcolor{gray!6}{0.2467} & \cellcolor{gray!6}{0.2779}\\
\hline
\end{tabular}}
\end{table}

%% In the same way that the tables and graphs are presented comparing with the highest score, it could be compared with any other team, or even with the baseline (if it exists). In the latter case, the teams that exceed the baseline could be identified. {\color{blue} ¿quitar?}

%% Just as performance metrics are analyzed through confidence intervals, bivariate confidence regions can be constructed for pairs of metrics to visualize and analyze the relationship of metrics for different competitors.

\subsection{Statistical Significance Testing}
\label{sec:significance}

In sub-section \ref{sec:independents}, ordered confidence intervals were shown, and in sub-section \ref{sec:paired}, comparisons were made with the best-performing competitor. These comparisons raise the question of whether the hypothesis of equality versus difference should be evaluated, given that it is evident that one competitor has better performance than the other in the test dataset. The previous question can be addressed by comparing the performance of two competitors, $A$ and $B$, to determine whether $A$ is superior to $B$ in a large population of data, i.e., $\theta_A > \theta_B$. Given the test dataset $x=x_1, \ldots , x_n$, assume that $A$ beats $B$ by a magnitude $\delta(x)=\theta_A (x) - \theta_B (x)$, the null hypothesis, $H_0$, is that $A$ is not superior to $B$ in the total population, and $H_1$ is that it is. Therefore, the aim is to determine how likely it would be for a similar victory for $A$ to occur in a new independent test dataset, denoted as $y$, assuming that $H_0$ is true.

Hypothesis testing aims to calculate the probability $p(\delta(X)>\delta(x) \mid H_0,x)$, where $X$ represents a random variable that considers the possible test sets of size $n$ that we could have selected, while $\delta(x)$ refers to the observed difference, that is, it is a constant. $p(\delta(X)>\delta(x) \mid H_0,x)$ is called the $p-value(x)$, and traditionally, if $p-value(x)<0.05$, the observed value $\delta(x)$ is considered sufficiently unlikely to reject $H_0$, meaning that the evidence suggests that $A$ is superior to $B$, see \cite{Berg-Kirkpatrick2012}.

In most cases, the $p-value(x)$ is not easily calculated and must be approximated. As described, this work uses paired bootstrap, not only because it is the most widely used (see \cite{Berg-Kirkpatrick2012,Bisani2004,Zhang2004,Koehn2004}), but also because it can be easily applied to any performance metric.

\begin{figure}[ht]
\centering
\includegraphics[width=1\textwidth]{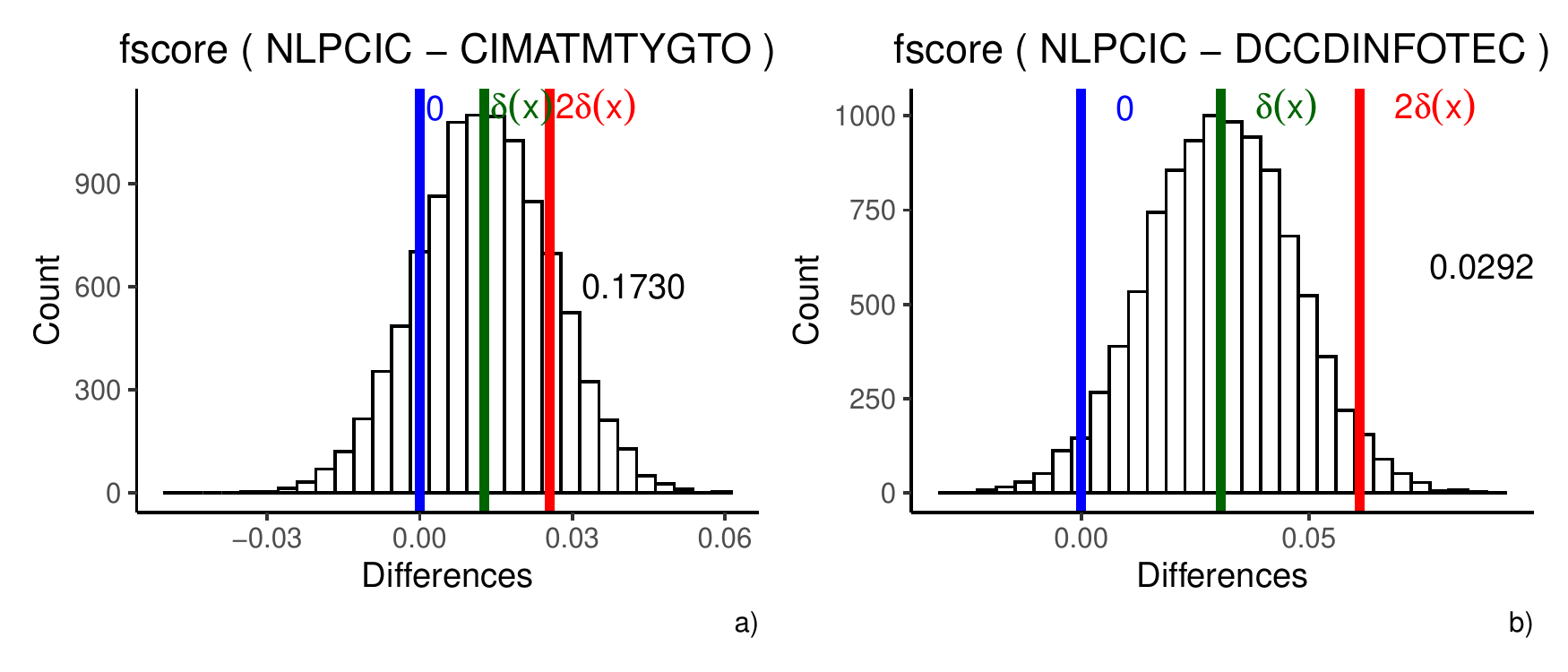}
\caption{ Bootstrap distribution of the $F_1$ score differences between  $NLPCIC$ and $CIMATMTYGTO$ (a), and $NLPCIC$ and $DCCDINFOTEC$ (b)}
\label{fig:histogram}
\end{figure}

As shown in \cite{Berg-Kirkpatrick2012}, the $p-value(x)$ can be estimated by computing the fraction of times that this difference is greater than $2\delta(x)$. It is crucial to keep in mind that this distribution is centered around $\delta(x)$, given that $X$ is drawn from $x$, where it is observed that $A$ is superior to $B$ by $\delta(x)$. Figure \ref{fig:histogram} illustrates the $p-value(x)$ process by showing the bootstrap distribution of the $F_1$ score differences between $NLPCIC$ and $CIMATMTYGTO$ (a), and $NLPCIC$ and $DCCDINFOTEC$ (b). The values zero, $\delta(x)$, and $2\delta(x)$ are highlighted for better understanding. When comparing $NLPCIC$ and $CIMATMTYGTO$ in the test dataset $x$, the difference $\delta(x)=0.7154-0.7026=0.0128$, is not significant at the 5\% level because the $p-value(x)$ is $0.1730$. On the other hand, when comparing $NLPCIC$ and  $DCCDINFOTEC$, $\delta(x)=0.7154-0.6847=0.0307$, which is significant at the 5\% level with a $p-value(x)$ of $0.0292$. In other words, $NLPCIC$ is not significantly better than $CIMATMTYGTO$ but better than $DCCDINFOTEC$. In section \ref{sec:paired}, it was shown through confidence intervals that the evidence supports $H_0$ (same performance) instead of $H_1$ (difference in performance). If we estimate the $p-value(x)$, it would be approximately $2\times 0.0292=0.0584$, which is not significant at the 5\% level. Table \ref{tab:fscoretab} summarizes the differences in the $F_1$ score and their significance. One can observe, for instance, that $NLPCIC$ is better than $CIMATGTO$ by $0.036$ with a $1\%$ significance.

\begin{table}[!h]
\caption{\label{tab:fscoretab} Differences of $F_1$ score (column)-(row), and their significance.  \\ Note: $\dagger p<.1$, {\footnotesize*}$p<.05$,
{\footnotesize**}$p<.1$,  and
{\footnotesize***}$p<.001$.}
\centering
\resizebox{\linewidth}{!}{
\fontsize{7}{9}\selectfont
\begin{tabular}[t]{llllllllll}
\hline
  & NLPCIC & CIMATMTYGTO & DCCDINFOTEC & CIMATGTO & UMUTeam & Timen & CICIPN & xjywing & aomar\\
\hline
\cellcolor{gray!6}{CIMATMTYGTO} & \cellcolor{gray!6}{0.013} & \cellcolor{gray!6}{} & \cellcolor{gray!6}{} & \cellcolor{gray!6}{} & \cellcolor{gray!6}{} & \cellcolor{gray!6}{} & \cellcolor{gray!6}{} & \cellcolor{gray!6}{} & \cellcolor{gray!6}{}\\
DCCDINFOTEC & 0.031 * & 0.018 &  &  &  &  &  &  & \\
\cellcolor{gray!6}{CIMATGTO} & \cellcolor{gray!6}{0.036 **} & \cellcolor{gray!6}{0.023 *} & \cellcolor{gray!6}{0.006} & \cellcolor{gray!6}{} & \cellcolor{gray!6}{} & \cellcolor{gray!6}{} & \cellcolor{gray!6}{} & \cellcolor{gray!6}{} & \cellcolor{gray!6}{}\\
UMUTeam & 0.045 ** & 0.032 ** & 0.014 & 0.009 &  &  &  &  & \\
\cellcolor{gray!6}{Timen} & \cellcolor{gray!6}{0.111 ***} & \cellcolor{gray!6}{0.099 ***} & \cellcolor{gray!6}{0.081 ***} & \cellcolor{gray!6}{0.075 ***} & \cellcolor{gray!6}{0.067 ***} & \cellcolor{gray!6}{} & \cellcolor{gray!6}{} & \cellcolor{gray!6}{} & \cellcolor{gray!6}{}\\
CICIPN & 0.114 *** & 0.101 *** & 0.083 *** & 0.077 *** & 0.069 *** & 0.002 &  &  & \\
\cellcolor{gray!6}{xjywing} & \cellcolor{gray!6}{0.222 ***} & \cellcolor{gray!6}{0.209 ***} & \cellcolor{gray!6}{0.191 ***} & \cellcolor{gray!6}{0.185 ***} & \cellcolor{gray!6}{0.177 ***} & \cellcolor{gray!6}{0.110 ***} & \cellcolor{gray!6}{0.108 ***} & \cellcolor{gray!6}{} & \cellcolor{gray!6}{}\\
aomar & 0.242 *** & 0.230 *** & 0.212 *** & 0.206 *** & 0.198 *** & 0.131 *** & 0.129 *** & 0.021 *** & \\
\cellcolor{gray!6}{CENAmrita} & \cellcolor{gray!6}{0.247 ***} & \cellcolor{gray!6}{0.234 ***} & \cellcolor{gray!6}{0.216 ***} & \cellcolor{gray!6}{0.211 ***} & \cellcolor{gray!6}{0.202 ***} & \cellcolor{gray!6}{0.135 ***} & \cellcolor{gray!6}{0.133 ***} & \cellcolor{gray!6}{0.025 ***} & \cellcolor{gray!6}{0.004}\\
\hline
\end{tabular}}
\end{table}

\section{Conclusions}
\label{sec:conclusions}

This paper has presented a procedure that uses bootstrap to infer the performance of classifiers from different teams in a competition. Confidence intervals were provided for each competitor and for differences in performance compared with the best-performing competitor. By graphing these confidence intervals, it can be quickly determined whether the differences are significant or not. The significance calculation was also presented to compare whether one competitor is better. It was highlighted that these ideas can be easily applied to any classification or regression problem. In summary, these proposals offer valuable tools for challenge organizers when making decisions.

\bibliographystyle{abbrv}
\bibliography{mybibliography}
\end{document}